\documentclass[11pt]{article}

\usepackage[final]{acl}

\usepackage{times}
\usepackage{latexsym}

\usepackage[T1]{fontenc}
\usepackage[utf8]{inputenc}

\usepackage{microtype}

\usepackage{inconsolata}

\usepackage{graphicx}

\usepackage{amsmath}
\usepackage{float}
\usepackage{pifont}
\newcommand{\cmark}{\ding{51}} 
\newcommand{\xmark}{\ding{55}} 
\usepackage{amssymb}

\usepackage{booktabs}      % \toprule \midrule \bottomrule
\usepackage{makecell}      % \makecell
\usepackage{tabularx}      % tabularx + X column
\usepackage{array}         % \arraybackslash
\usepackage[table]{xcolor} % \rowcolor + \textcolor

\usepackage{soul}          % \hl \sethlcolor
\definecolor{lightred}{RGB}{255,220,220}
\definecolor{lightyellow}{RGB}{255,245,204}

\providecommand{\greencheck}{\textcolor{green}{\cmark}}

\usepackage{algorithm}
\usepackage{algorithmic}
\usepackage{enumitem}
\usepackage{mdframed} % 可选：想要更像代码框就用
\setlist[itemize]{leftmargin=*, nosep, topsep=2pt}

\usepackage{caption}
\usepackage{hyperref}
\hypersetup{hidelinks}

\usepackage{tcolorbox}
\tcbset{
  colback=white,
  colframe=black!55,
  boxrule=0.35pt,
  arc=1.2pt,
  left=6pt,
  right=6pt,
  top=4pt,
  bottom=4pt
}

\usepackage{fvextra}
\usepackage{xcolor}

\DefineVerbatimEnvironment{promptblock}{Verbatim}{
  fontsize=\small,
  formatcom=\ttfamily,
  xleftmargin=1.2em,
  breaklines=true,
  breakanywhere=true,
  commandchars=\\\{\}
}

\usepackage{pgfplots}
\pgfplotsset{compat=1.17}

\pgfplotsset{
    /pgfplots/ybar legend/.style={
        /pgfplots/legend image code/.code={%
            \draw[#1,/tikz/.cd,yshift=-0.25em]
            (0cm,0cm) rectangle (1em,0.8em);},
    },
}

\title{SEMA-RAG: A Self-Evolving Multi-Agent Retrieval-Augmented Generation Framework for Medical Reasoning}

\author{
Yongfeng Huang\textsuperscript{1}\thanks{These authors contributed equally to this work.}
\quad
Ruiying Chen\textsuperscript{2}\footnotemark[1]
\quad
James Cheng\textsuperscript{1}\thanks{Corresponding author.}
\\
\textsuperscript{1}CSE, The Chinese University of Hong Kong
\quad
\textsuperscript{2}Wuhan University of Technology
\\
\texttt{\{yfhuang22,jcheng\}@cse.cuhk.edu.hk}
\quad
\texttt{355227@whut.edu.cn}
}

\begin{document}
\maketitle
\begin{abstract}
Retrieval-Augmented Generation (RAG) is widely employed to mitigate risks such as hallucinations and knowledge obsolescence in medical question answering, yet its predominantly single-round, static retrieval paradigm misaligns with the multi-stage process of clinical reasoning. This compressed workflow induces two structural deficiencies: question-to-query translation often lacks clinically grounded semantic interpretation, and retrieval lacks iterative sufficiency feedback, making it difficult to form reliable evidence chains. We argue that both issues stem from a deeper cause—overloading a single reasoning chain with heterogeneous tasks of interpretation, exploration, and adjudication—and that the remedy is to reconstruct the workflow via task decoupling and dynamic multi-round exploration. 
To this end, we propose \textbf{SEMA-RAG}, a \textbf{S}elf-\textbf{E}volving \textbf{M}ulti-\textbf{A}gent RAG framework for medical question answering, which assigns these roles to three specialist agents: the \textbf{Interpreter Agent} for clinical schema interpretation, the \textbf{Explorer Agent} for sufficiency-driven self-evolving retrieval, and the \textbf{Arbiter Agent} for evidence adjudication and answer selection. Across five benchmarks and five LLM backbones, SEMA-RAG improves the strongest baseline by \textbf{+6.46} accuracy points on average, measured per backbone.
\end{abstract}

\section{Introduction}

In recent years, large language models (LLMs) have shown specific capabilities in understanding and reasoning about medical knowledge when applied in healthcare \cite{kung_performance_2023,omar_large_2024}. However, they remain prone to hallucinations and outdated information in high-stakes clinical settings \cite{omiye_large_2024,roustan_clinicians_2025}. Retrieval-Augmented Generation (RAG), which incorporates external authoritative evidence to support the generation process, has been widely adopted to mitigate these risks \cite{10.5555/3495724.3496517}.

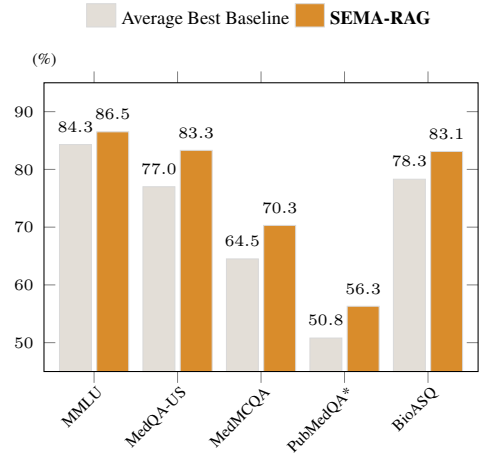
\begin{figure}[H]
\centering
% 定义颜色
\definecolor{bancolor}{HTML}{E3DED7}
\definecolor{semacolor}{HTML}{D98C2B}

\begin{tikzpicture}
    \begin{axis}[
        ybar,
        axis on top,
        set layers=standard,
        width=0.95\columnwidth,
        height=5.4cm,
        bar width=12pt,
        % --- 1. 调整 Y 轴范围和刻度 ---
        ymin=45, ymax=95,
        ytick={50, 60, 70, 80, 90}, % <--- 显式加入 70
        % ---------------------------
        ylabel={},
        clip=false,
        axis line style={draw=black},
        y axis line style={draw=black},
        x axis line style={draw=black},
        % --- 2. 缩小坐标轴数字字体 ---
        tick label style={
            color=black, 
            font=\tiny       % <--- 改为 \tiny，让纵坐标数字变小
        },
        % ---------------------------
        symbolic x coords={MMLU, MedQA-US, MedMCQA, PubMedQA*, BioASQ},
        xtick=data,
        ymajorgrids=false,
        grid style={solid, gray!15},
        nodes near coords,
        nodes near coords style={
            font=\tiny\bfseries, % 柱子顶上的数字保持极小加粗
            color=black,
            yshift=1pt,
            /pgf/number format/.cd,
            fixed, fixed zerofill, precision=1
        },
        xticklabel style={
            font=\tiny,      % <--- X轴标签也同步改为 \tiny 以保持协调
            rotate=45,           
            anchor=north east,   
            inner sep=0pt,       
            yshift=-1pt          
        },
        legend style={
            at={(0.5, 1.15)},    
            anchor=south,
            legend columns=-1,
            draw=none,
            fill=none,
            font=\scriptsize
        },
        enlarge x limits=0.15,
    ]
        % Baseline 数据
        \addplot[
            fill=bancolor,
            draw=black!15,
            line width=0.3pt
        ] coordinates {
            (MMLU, 84.3) (MedQA-US, 77.0) (MedMCQA, 64.5) (PubMedQA*, 50.8) (BioASQ, 78.3)
        };
        
        % SEMA-RAG 数据
        \addplot[
            fill=semacolor,
            draw=black!15,
            line width=0.3pt
        ] coordinates {
            (MMLU, 86.5) (MedQA-US, 83.3) (MedMCQA, 70.3) (PubMedQA*, 56.3) (BioASQ, 83.1)
        };
        
        \legend{Average Best Baseline, \textbf{SEMA-RAG}}
        
        % 单位标签
        \node[anchor=south west, font=\tiny, color=black] 
        at (rel axis cs:-0.05, 1.02) {(\%)};
    \end{axis}
\end{tikzpicture}
\vspace{-0.5em}
\caption{Benchmark-level accuracy averaged over five LLM backbones.}
\label{fig:intro_performance_final}
\end{figure}

However, standard RAG frameworks typically treat retrieval as a static, single-round auxiliary step, misaligning with the multi-stage process of clinical reasoning: clinicians often first interpret patient narratives as searchable clinical questions, then progressively gather and verify information to address evidence gaps, and weigh and integrate redundant or contradictory evidence to ultimately form judgments based on relatively robust evidence \cite{linn2012clinicalreasoning,yazdani_models_2017}. In contrast, single-round static RAG compresses this process into a single retrieval and generation step. This is akin to requiring clinicians to simultaneously analyze, retrieve, evaluate, and diagnose immediately upon receiving initial medical records, without adjusting their reasoning as new evidence emerges. This typically leads to two structural flaws: \emph{(i)} \textbf{the translation from question to query lacks clinical semantic interpretation}, making implicit constraints difficult to articulate explicitly \cite{10.1002/asi.23924}; and \emph{(ii)} \textbf{the retrieval process lacks mechanisms for sufficiency assessment and feedback}, hindering self-evolving iterative convergence under insufficient evidence and thus weakening reliable evidence-chain formation \cite{mallen-etal-2023-trust,10.5555/3618408.3619699}.

% Further analysis suggests that these shortcomings arise not from insufficient knowledge, but from overloading heterogeneous tasks into a single reasoning chain. 
These shortcomings, we argue, are not independent issues but rather symptoms of a deeper problem: \textbf{overloading heterogeneous tasks into a single reasoning chain}.
When question interpretation, evidence exploration, and answer adjudication are tightly coupled, the cognitive load increases and the steps become interdependent, making it hard for the model to promptly adjust retrieval and reasoning when evidence is insufficient or conflicting \cite{wang-etal-2023-plan,liu-etal-2024-lost}. Thus, the key is not to intensify single-round reasoning, but to restructure RAG to better match the phased clinical workflow by extending single-round queries into multi-round iterative exploration. After each retrieval round, the system evaluates whether the evidence covers key constraints and then chooses the next action: terminate exploration and proceed to decision integration if sufficient, or generate targeted follow-up queries to fill gaps if insufficient. 
This mechanism, which continuously updates the direction of queries and retrieval based on the evaluation results of each round, enables the system to adjust and converge as evidence accumulates progressively. In this sense, the process constitutes a form of \textbf{intra-test-time self-evolution}, in which the system adaptively updates its query and retrieval trajectory during task execution while remaining tightly coupled to the current problem instance \cite{gao_survey_2026}. For simplicity, we use the term "self-evolving" in the remainder of the paper to refer to this intra-test-time setting.

To this end, we propose \textbf{SEMA-RAG} (\textbf{S}elf-\textbf{E}volving \textbf{M}ulti-\textbf{A}gent RAG). This framework simulates clinical workflows through task decoupling and role specialization, decomposing complex clinical reasoning into three collaborative modules: \textbf{Interpreter Agent (I-Agent)} maps unstructured inputs to structured clinical semantics; \textbf{Explorer Agent (E-Agent)} implements self-evolving, evidence-sufficiency-driven retrieval for convergent exploration; \textbf{Arbiter Agent (A-Agent)} performs comprehensive adjudication based on closed-loop evidence.

We evaluate SEMA-RAG on five medical question-answering benchmarks.
As shown in Figure~\ref{fig:intro_performance_final}, SEMA-RAG consistently outperforms representative baselines in terms of average accuracy on each benchmark when averaged over multiple underlying LLMs. 
Across five benchmarks and five LLM backbones, it improves the strongest baseline by an average of \textbf{+6.46} accuracy points, validating convergent evidence-chain construction via task decoupling, role specialization, and evidence sufficiency-driven self-evolving retrieval. Our main contributions are as follows:

\begin{itemize}
\item We propose \textbf{SEMA-RAG}, a multi-agent RAG framework for medical question answering, which models clinical reasoning processes via role division and collaboration.
\item We develop a self-evolving \textbf{Explorer Agent} that updates queries based on evidence gaps, steering retrieval toward medical reasoning objectives.
\item We validate SEMA-RAG on five medical Q\&A benchmarks across multiple underlying LLMs, achieving consistent improvements over baselines.
\end{itemize}

\begin{figure*}[t]
  \centering
  \includegraphics[width=\textwidth]{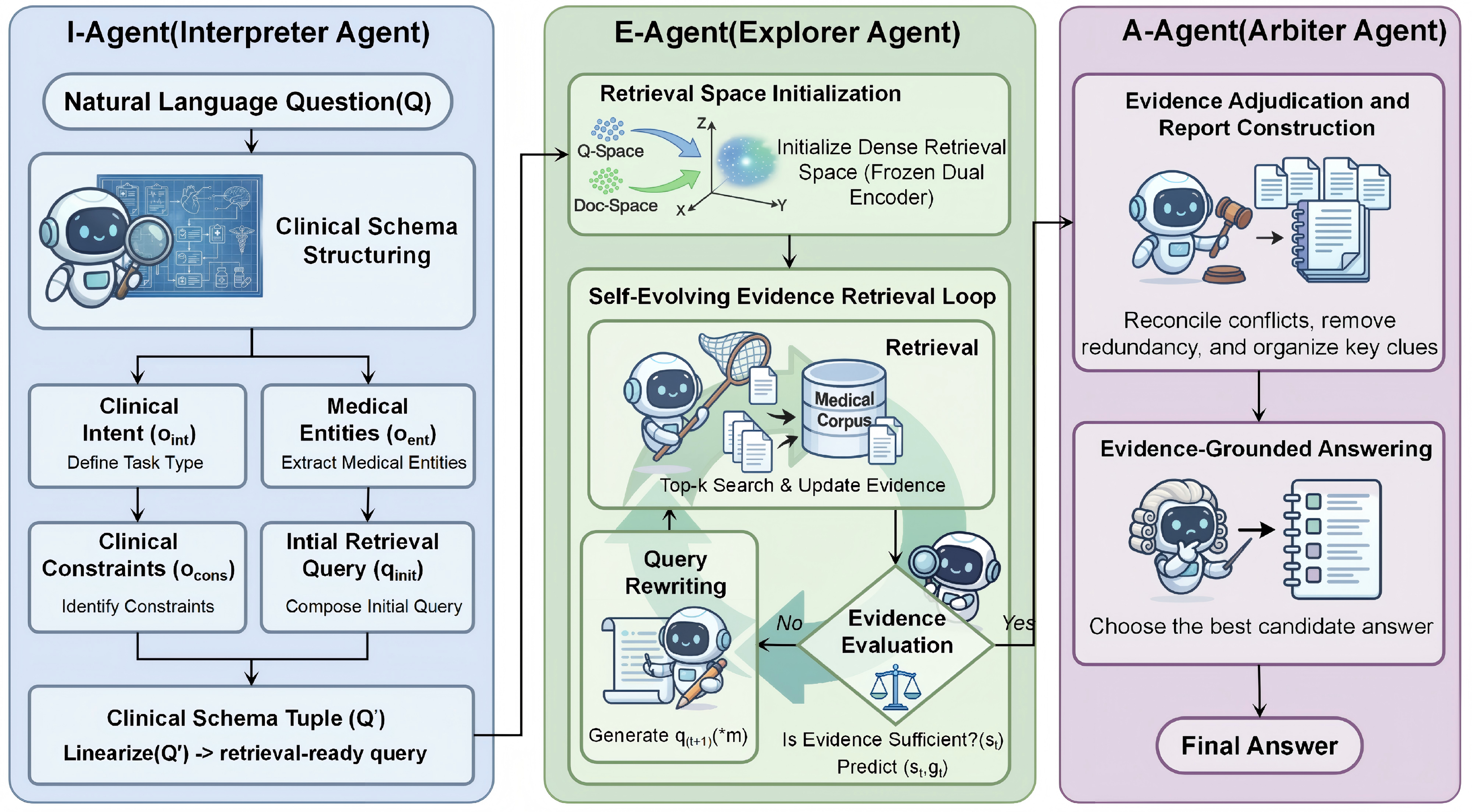}
  \caption{Overview of \textbf{SEMA-RAG}: (i) \textbf{I-Agent} structures the input question $Q$ into a clinical schema tuple $Q'$ for retrieval; (ii) \textbf{E-Agent} conducts sufficiency-driven self-evolving multi-round retrieval to obtain a converged evidence set $C^*$; (iii) \textbf{A-Agent} adjudicates evidence into a traceable report $R$ and selects the final answer grounded in $R$.}
  \label{fig:method}
\end{figure*}

\section{Preliminaries}

\subsection{Task Formulation of Medical RAG}

Given a medical question \(Q\), the system selects the final answer \(\tilde{y}\) from the discrete candidate set \(\mathcal{Y}\) (\(y \in \mathcal{Y}\)). Under question-only retrieval conditions, the system may only retrieve evidence from the medical corpus \(\mathcal{C}\). The core RAG consists of a retrieval operator $\mathrm{Ret}(\cdot)$ and a generation operator:
$C = \mathrm{Ret}(Q)$, and predicts accordingly:
\[
\tilde{y} = \arg\max_{y \in \mathcal{Y}} p(y \mid Q, C).
\]

\subsection{Multi-Agent Roles and Abstraction}
We employ role-based division of labor, with three agents collaborating to complete the medical question answering process: \textbf{I-Agent} handles question interpretation, \textbf{E-Agent} manages evidence exploration, and \textbf{A-Agent} oversees answer adjudication.

Three agents share the same underlying language model, differentiated solely by role-specific prompts. We denote the output of the shared LLM conditioned on a role prompt $Pmt_r$ and an input $X$ as $\mathrm{Agent}_r(Pmt_r,X)$, where $X$ may be a set containing multiple elements.

\section{Method}
Figure~\ref{fig:method} illustrates the overall SEMA-RAG framework, which comprises three role-based agents with responsibilities described below.

\subsection{I-Agent as a Question Interpreter}
I-Agent does not merely rephrase the input medical question \(Q\); instead, it semantically structures \(Q\) and projects it onto an explicit \textbf{Clinical Schema}. This process externalizes latent clinical intent and key constraints, providing stable anchors for subsequent retrieval and reasoning.

Specifically, I-Agent produces a clinical schema tuple \(Q'\) with four components:
\emph{(i)} clinical intent \(o_{\mathrm{int}}\), describing the implied task type (e.g., diagnosis, treatment, dosage);
\emph{(ii)} medical entities \(o_{\mathrm{ent}}\), identifying core medical objects (e.g., diseases, drugs);
\emph{(iii)} clinical constraints \(o_{\mathrm{cons}}\), specifying applicability conditions (e.g., pregnant, renal impairment, adult);
and \emph{(iv)} an initial retrieval query \(q_{\mathrm{init}}\), a concise, search-oriented question distilled from the above schema.

Formally, I-Agent maps \(Q\) to the schema tuple:
\[
\begin{aligned}
Q' &= \langle o_{\mathrm{int}},\, o_{\mathrm{ent}},\, o_{\mathrm{cons}},\, q_{\mathrm{init}}\rangle \\
   &= \mathrm{Agent}_{\mathrm{I}}(Pmt_{\mathrm{I}},\, Q).
\end{aligned}
\]

To make the schema tuple usable by the dense retriever, we further linearize \(Q'\) into a retrieval-ready query string:
\[
\hat{q}_{\mathrm{init}} = \mathrm{Linearize}(Q')
\]
\[
= \mathrm{Concat}(q_{\mathrm{init}}, \oplus, o_{\mathrm{int}}, \oplus, o_{\mathrm{ent}}, \oplus, o_{\mathrm{cons}}),
\]
where \(\oplus\) denotes a semicolon separator. Here, \(\mathrm{Linearize}(\cdot)\) is a parameter-free function without field-specific weights or additional control tokens. It preserves \(q_{\mathrm{init}}\) as the core query while explicitly incorporating \(o_{\mathrm{int}}, o_{\mathrm{ent}}\), and \(o_{\mathrm{cons}}\), making clinically important but implicit constraints more visible to the retriever and reducing semantic drift in the initial retrieval stage. The resulting query \(\hat{q}_{\mathrm{init}}\) is used to initialize E-Agent, while \(Q'\) remains the clinical anchor for subsequent coordination.

\subsection{E-Agent as a Knowledge Explorer}
E-Agent begins with the linearized schema query \(\hat{q}_{\mathrm{init}}\) generated by I-Agent and progressively completes the evidence through a self-evolving iterative retrieval process, ultimately constructing the final evidence set \(C^*\).

\paragraph{Retrieval Space Initialization}
We construct a dense vector retrieval space based on the medical corpus $\mathcal{C}$. Using a parameter-frozen medical dual encoder, we map queries and documents to the same vector space, where $E_{\mathrm{qry}}(\cdot)$ and $E_{\mathrm{doc}}(\cdot)$ denote the query and document encoders, respectively. Given a query $q$, its Top-$k$ candidate documents (passages/chunks) are retrieved based on vector similarity as follows:
\[
\mathrm{TopK}(q)
=
\operatorname*{Top\text{-}k}_{D \in \mathcal{C}}
\left\langle E_{\mathrm{qry}}(q),\, E_{\mathrm{doc}}(D) \right\rangle .
\]
where \(\operatorname*{Top\text{-}k}\) returns the \(k\) documents with the largest similarity scores.

\paragraph{Self-Evolving Evidence Retrieval Loop}
Using the linearized schema query $\hat{q}_{\mathrm{init}}$ as the initial query, we set $\mathcal{Q}_1=\{\hat{q}_{\mathrm{init}}\}$ and $C_0=\varnothing$, where $\mathcal{Q}_t$ is the query set for the $t$th retrieval round and $C_t$ is the accumulated evidence set after round $t$. Each retrieved document $D_i$ is associated with a deterministic document identifier $\mathrm{ID}(D_i)$, which is retained throughout the pipeline for exact deduplication and source tracing. At round $t$, E-Agent performs retrieval for each query in $\mathcal{Q}_t$ and updates the evidence set:
\[
\mathcal{D}_t=\bigcup_{q\in\mathcal{Q}_t}\mathrm{TopK}(q),
\]
\[
C_t=C_{t-1}\cup\{D_i\in\mathcal{D}_t : \mathrm{ID}(D_i)\notin \mathrm{IDs}(C_{t-1})\}.
\]

Conditioned on clinical anchors $Q'$, the current textual query set $\mathcal{Q}_t$, and the evidence set $C_t$, E-Agent predicts a sufficiency flag $s_t$, a gap description $g_t$, and the next query set $\mathcal{Q}_{t+1}$:
\[
\begin{aligned}
[s_t,\, g_t,\, \mathcal{Q}_{t+1}]
&=
\mathrm{Agent}_{\mathrm{E}}\!\left(Pmt_{\mathrm{E}},\, [Q',\, \mathcal{Q}_t,\, C_t]\right), \\
\mathcal{Q}_{t+1}
&=
\left\{q_{t+1}^{\langle 1\rangle},\ldots,q_{t+1}^{\langle m\rangle}\right\}.
\end{aligned}
\]
where $s_t\in\{0,1\}$ indicates evidence sufficiency: if $s_t=1$, the evidence is sufficient and we set $\mathcal{Q}_{t+1}=\varnothing$; otherwise ($s_t=0$), evidence gaps remain, and $g_t$ identifies missing conditions or reasoning steps, from which $\mathcal{Q}_{t+1}$ generates $m$ candidate follow-up queries targeting these gaps.

When $s_t=0$, the generated $\mathcal{Q}_{t+1}$ is issued in the next round to retrieve additional evidence, and the results are incorporated into the update of $C_{t+1}$.

Iteration terminates when $s_t=1$, $t=T_{\max}$, or stagnation occurs (i.e., $\mathcal{Q}_{t+1}=\varnothing$). Upon termination, we obtain the closed evidence set, record the actual number of iterations $T\le T_{\max}$, and store the self-evolving trajectory:
\[
C^* = C_T,
\;
\tau=\left\{[\mathcal{Q}_1, C_1],\ldots,[\mathcal{Q}_T, C_T]\right\}.
\]

\subsection{A-Agent as an Evidence Arbiter}
A-Agent adjudicates evidence by organizing the converged set 
\(C^*\) into a traceable evidence report and generating a discrete answer from it.

\paragraph{Evidence Adjudication and Report Construction}
Given redundant and potentially conflicting evidence, A-Agent first adjudicates $C^*$ by removing irrelevant or duplicated content, identifying consistencies and conflicts, and organizing supporting and refuting clues into a structured evidence report $R$. For traceability, we retain the original document identifier for each retrieved document $D_i\in C^*$, forming the source set 
\[
\mathcal{S}^*=\{(\mathrm{ID}(D_i), D_i)\mid D_i\in C^*\}.
\]
A-Agent then generates the evidence report
\[
R=\mathrm{Agent}_{\mathrm{A}}(Pmt_{\mathrm{adj}}, [Q, C^*, \mathcal{S}^*]),
\]
where \(R\) explicitly organizes key conclusions relevant to the question along with their source indices, provides a reconciled synthesis of conflicting evidence, and offers a stable basis for final answer selection.

\paragraph{Evidence-Grounded Answering}
Upon obtaining the evidence report \(R\), A-Agent performs discrete answer selection over the candidate answer set \(\mathcal{Y}\):
\[
\tilde{y} = \mathrm{Agent}_{\mathrm{A}}( Pmt_{\mathrm{ans}}, [Q, R]),
\]
where \(\tilde{y}\) is the final predicted answer.

\section{Experiments}

\subsection{Experimental Setup}

\begin{table*}[t]
\centering
\resizebox{\textwidth}{!}{%
\begin{tabular}{lcccccc}
\hline
\textbf{Model} & \textbf{MMLU-Med} & \textbf{MedQA-US} & \textbf{MedMCQA} & \textbf{PubMedQA*} & \textbf{BioASQ-Y/N} & \textbf{Average} \\
\hline

% --- Deepseek-v3.1 Block ---
\multicolumn{7}{l}{\underline{\textbf{\texttt{deepseek-v3.1}}} \cite{deepseek-ai_deepseek-v3_2025}} \\
\quad + CoT \cite{10.5555/3600270.3602070}           & 88.15 & \underline{77.53} & \underline{71.69} & 38.40 & 80.10 & 71.17 \\
\quad + MedCPT \cite{jin_medcpt_2023}                & 85.12 & 73.84 & 62.66 & 43.20 & 76.38 & 68.24 \\
\quad + MedRAG \cite{xiong-etal-2024-benchmarking}   & \underline{88.61} & 77.14 & 67.99 & 44.60 & 78.48 & 71.36 \\
\quad + \textit{i}-MedRAG \cite{xiong_improving_2025}         & 85.86 & 74.78 & 65.65 & \underline{50.60} & \underline{80.58} & \underline{71.49} \\
\quad + SEMA-RAG (Ours)                             & \textbf{91.46} & \textbf{89.95} & \textbf{75.09} & \textbf{59.20} & \textbf{82.85} & \textbf{79.71} \\
\hline

% --- Kimi-k2 Block ---
\multicolumn{7}{l}{\underline{\textbf{\texttt{kimi-k2}}} \cite{team_kimi_2025}} \\
\quad + CoT \cite{10.5555/3600270.3602070}           & 84.39 & 77.85 & 72.08 & 53.60 & \underline{85.76} & 74.74 \\
\quad + MedCPT \cite{jin_medcpt_2023}                & 89.81 & 80.68 & 73.85 & 50.20 & 81.39 & 75.19 \\
\quad + MedRAG \cite{xiong-etal-2024-benchmarking}   & \underline{91.37} & 81.54 & 73.20 & 52.60 & 85.60 & 76.86 \\
\quad + \textit{i}-MedRAG \cite{xiong_improving_2025}         & 91.28 & \underline{81.78} & \underline{74.13} & \underline{54.60} & 83.17 & \underline{76.99} \\
\quad + SEMA-RAG (Ours)                             & \textbf{91.46} & \textbf{86.41} & \textbf{76.07} & \textbf{55.80} & \textbf{88.67} & \textbf{79.68} \\
\hline

% --- Qwen3-coder-plus Block ---
\multicolumn{7}{l}{\underline{\textbf{\texttt{qwen3-coder-plus}}} \cite{yang_qwen3_2025}} \\
\quad + CoT \cite{10.5555/3600270.3602070}           & 89.26 & 76.90 & \underline{73.06} & 47.20 & 81.72 & \underline{73.63} \\
\quad + MedCPT \cite{jin_medcpt_2023}                & 87.42 & 75.26 & 67.44 & 46.60 & 75.89 & 70.52 \\
\quad + MedRAG \cite{xiong-etal-2024-benchmarking}   & \underline{89.44} & \underline{81.54} & 69.26 & \underline{49.20} & 72.33 & 72.35 \\
\quad + \textit{i}-MedRAG \cite{xiong_improving_2025}         & 89.26 & 77.38 & 70.26 & 48.60 & \underline{82.52} & 73.60 \\
\quad + SEMA-RAG (Ours)                             & \textbf{92.10} & \textbf{86.17} & \textbf{74.23} & \textbf{56.00} & \textbf{83.01} & \textbf{78.30} \\
\hline

% --- Gemini-2.0-flash Block ---
\multicolumn{7}{l}{\underline{\textbf{\texttt{gemini-2.0-flash}}} \cite{google_gemini2flash_modelcard_2025}} \\
\quad + CoT \cite{10.5555/3600270.3602070}           & 58.22 & 65.12 & 41.33 & 40.20 & 68.45 & 54.66 \\
\quad + MedCPT \cite{jin_medcpt_2023}                & 62.35 & 70.54 & 44.90 & 42.80 & 70.06 & 58.13 \\
\quad + MedRAG \cite{xiong-etal-2024-benchmarking}   & \underline{74.29} & \underline{83.19} & \underline{50.87} & 44.20 & 72.65 & \underline{65.04} \\
\quad + \textit{i}-MedRAG \cite{xiong_improving_2025}         & 65.47 & 77.69 & 46.78 & \underline{51.20} & \underline{77.99} & 63.83 \\
\quad + SEMA-RAG (Ours)                             & \textbf{80.99} & \textbf{90.42} & \textbf{71.60} & \textbf{59.20} & \textbf{88.19} & \textbf{78.08} \\
\hline

% --- GLM-4.0-flash Block ---
\multicolumn{7}{l}{\underline{\textbf{\texttt{glm-4.0-flash}}} \cite{glm_chatglm_2024}} \\
\quad + CoT \cite{10.5555/3600270.3602070}           & 68.14 & 50.43 & 48.22 & 40.00 & 58.90 & 53.14 \\
\quad + MedCPT \cite{jin_medcpt_2023}                & 73.00 & 58.21 & 50.63 & 42.20 & 60.84 & 56.98 \\
\quad + MedRAG \cite{xiong-etal-2024-benchmarking}   & \textbf{77.59} & \underline{60.88} & \underline{52.59} & 46.80 & 62.78 & \underline{60.13} \\
\quad + \textit{i}-MedRAG \cite{xiong_improving_2025}         & 73.28 & 53.57 & 51.47 & \underline{48.40} & \underline{64.72} & 58.29 \\
\quad + SEMA-RAG (Ours)                             & \underline{76.68} & \textbf{63.79} & \textbf{54.34} & \textbf{51.20} & \textbf{72.98} & \textbf{63.80} \\
\hline

\end{tabular}%
}
\caption{Accuracy (\%) comparison of \textbf{SEMA-RAG} and baselines on five medical QA benchmarks across different LLMs. Bold indicates the best result within each model block and underline indicates the second-best.}
\label{tab:main_results}
\end{table*}

\subsubsection{Evaluation Benchmarks}
To systematically evaluate SEMA-RAG's performance and generalisation capabilities across diverse medical question-answering scenarios, we select five widely used datasets from the MIRAGE benchmark \cite{xiong-etal-2024-benchmarking}: three medical examination datasets (MMLU-Med \cite{hendrycks_measuring_2021}, MedQA-US \cite{jin_what_2020}, MedMCQA \cite{pal_medmcqa_2022}) and two biomedical research QA datasets (PubMedQA* \cite{jin_pubmedqa_2019} and BioASQ-Y/N \cite{tsatsaronis_overview_2015,krithara_bioasq-qa_2023}). Together, they cover general medical knowledge, clinical examinations, and biomedical literature inference. Following MIRAGE’s filtering and preprocessing pipeline, we retain only discrete biomedical classification questions, use PubMedQA* (with the original evidence context removed), and apply question-only retrieval for all tasks.

\subsubsection{Models and Baselines}

To assess SEMA-RAG’s robustness across backbones, we instantiate the framework on five publicly accessible LLMs: \textbf{deepseek-v3.1} \cite{deepseek-ai_deepseek-v3_2025}, \textbf{kimi-k2} \cite{team_kimi_2025}, \textbf{qwen3-coder-plus} \cite{yang_qwen3_2025}, \textbf{gemini-2.0-flash} \cite{google_gemini2flash_modelcard_2025}, and \textbf{glm-4.0-flash} \cite{glm_chatglm_2024}. These models originate from different providers, encompass diverse pretraining configurations and capability focuses.

We selected three representative methods for comparison to characterize performance differences across no retrieval, single-round retrieval, and iterative retrieval. The no-retrieval setting employs \textbf{CoT} \cite{10.5555/3600270.3602070}, relying solely on model-internal knowledge for chain-of-reasoning. The single-round retrieval setting employs \textbf{MedCPT} \cite{jin_medcpt_2023} as the medical domain retriever, further contrasting it with \textbf{MedRAG} \cite{xiong-etal-2024-benchmarking}'s retrieval-fusion framework. The iterative retrieval setting utilizes \textbf{\textit{i}-MedRAG} \cite{xiong_improving_2025}, which generates subsequent queries to drive multi-round retrieval and accumulate evidence.

\subsubsection{Implementation Details}
Following \textit{i}-MedRAG \cite{xiong_improving_2025} and MedRAG \cite{xiong-etal-2024-benchmarking}, we retrieve from Textbooks \cite{app11146421} and StatPearls \cite{statpearls} on all benchmarks. We use MedCPT \cite{jin_medcpt_2023} as the dense retriever and perform FAISS-based retrieval over these corpora.

All methods are evaluated in a zero-shot setting. Unless stated otherwise, SEMA-RAG uses $T_{\max}=2$, $k=16$, and $m=3$. We set temperature to 1.0 for I/E-Agent and 0.0 for A-Agent. Other baselines follow their official settings.

\subsection{Main Results: Consistent and Significant Improvements}
Table~\ref{tab:main_results} reports results on five medical QA benchmarks across five underlying LLMs. Overall, \textbf{SEMA-RAG achieves the best average accuracy within every model block}, indicating that the improvement is backbone-agnostic rather than tied to a particular LLM.

A consistent pattern is that the gains are larger on benchmarks where premature commitment under incomplete evidence is costly. This matches SEMA-RAG’s self-evolving exploration: it checks evidence sufficiency, performs targeted follow-up retrieval when needed, and only then moves to adjudication, leading to more reliable evidence chains.

Overall, this consistent and substantial advantage is not accidental.
It directly stems from SEMA-RAG’s successful simulation of expert clinical reasoning by decoupling interpretation, exploration, and adjudication, thereby alleviating the cognitive overload bottleneck diagnosed in the Introduction.
To unpack the source of these gains, we next present a core component analysis.

\subsection{Core Component Analysis}

In this section, we quantitatively evaluate the contributions of the three roles in SEMA-RAG via role-wise removal analysis. Specifically, we compare the full framework with three variants: (i) \textbf{w/o I-Agent}, which removes the question interpretation module and directly uses the raw question as the initial retrieval query; (ii) \textbf{w/o E-Agent}, which removes the self-evolving retrieval mechanism and performs only a single round of static retrieval based on the linearized schema query generated by I-Agent; and (iii) \textbf{w/o A-Agent}, which removes the answer adjudication module and directly generates answers from the final evidence set. The results are summarized in Table~\ref{tab:ablation}.

\begin{table}[H]
\centering
\small
\setlength{\tabcolsep}{3pt}
\renewcommand{\arraystretch}{1.2} 
\begin{tabular}{ccc|cc}
\hline
\textbf{I-Agent} & \textbf{E-Agent} & \textbf{A-Agent} & \textbf{MedQA-US} & \textbf{PubMedQA*} \\
\hline
\xmark & \cmark & \cmark & 85.47 & 54.20 \\
\cmark & \xmark & \cmark & 83.58 & 50.80 \\
\cmark & \cmark & \xmark & 86.49 & 53.60 \\
\textbf{\cmark} & \textbf{\cmark} & \textbf{\cmark} & \textbf{89.95} & \textbf{59.20} \\
\hline
\end{tabular}
\caption{Role-wise removal results of SEMA-RAG on MedQA-US and PubMedQA* (deepseek-v3.1).}
\label{tab:ablation}
\end{table}

\subsubsection{Resolving Ambiguous Queries with I-Agent}
Table~\ref{tab:ablation} shows that removing the I-Agent consistently degrades performance, indicating that question-to-query translation benefits from clinically grounded semantic interpretation.
Without this step, queries often remain underspecified and fail to surface implicit constraints, which increases the chance that retrieval drifts toward generic, symptom-level evidence rather than the decision-critical clinical setting.

As shown in Table~\ref{tab:case_study} (Step~1), I-Agent makes \emph{hospital day 7} retrieval-actionable, anchoring evidence in the late-onset inpatient context and enabling option-level discrimination for \textit{S.\ aureus}.
By extracting a structured clinical schema and making such key conditions retrievable, I-Agent keeps retrieval aligned with the intended clinical scenario and provides a cleaner substrate for subsequent exploration and adjudication.

\subsubsection{Achieving Dynamic Reasoning with E-Agent}
Table~\ref{tab:ablation} shows that removing E-Agent causes the largest performance drop, highlighting that the core gain comes from a self-evolving, sufficiency-driven closed-loop retrieval rather than static retrieval. 
Without E-Agent, the system loses explicit sufficiency feedback and is more prone to stop with partially covered evidence, leaving key constraints unresolved.

\begin{figure}[H]
    \centering
    % 1. 重新定义颜色，确保不丢失
    \definecolor{semacolor}{HTML}{D98C2B}
    
    \begin{tikzpicture}
        \begin{axis}[
            width=0.9\columnwidth, 
            height=5.5cm,
            xlabel={Max Iterations ($T_{\max}$)},
            ylabel={Accuracy (\%)},
            clip=false,
            xmin=0.5, xmax=9.5,
            ymin=86, ymax=91,
            xtick={1, 2, 3, 5, 7, 9},
            ytick={86, 87, 88, 89, 90, 91},
            ymajorgrids=true,
            grid style={dashed, gray!30},
            axis line style={draw=black!60},
            tick label style={font=\footnotesize}, % 坐标轴刻度字体
            label style={font=\small},             % 坐标轴名称字体
            legend style={at={(0.95,0.05)}, anchor=south east, font=\scriptsize, draw=none}
        ]
            \addplot[
                color=semacolor,       % 线条颜色
                mark=*,
                line width=1.2pt,
                mark size=1.8pt,
                nodes near coords, 
                % --- 核心修改：让数字变小 ---
                nodes near coords style={
                    font=\tiny,        % 设定为极小字体
                    scale=0.8,         % 再次强制缩放到 80%，确保变小
                    color=black,       % 数字颜色（黑色更清晰）
                    yshift=3pt,        % 稍微贴近点
                    /pgf/number format/fixed,
                    /pgf/number format/fixed zerofill,
                    /pgf/number format/precision=2
                }
            ] coordinates {
                (1, 86.57) 
                (2, 89.95) 
                (3, 90.10) 
                (5, 89.87) 
                (7, 88.45) 
                (9, 87.67)
            };
            \legend{SEMA-RAG ($m=3$)}
            
            % 虚线辅助线
            \draw[gray, dashed, opacity=0.7] (axis cs:2, 86) -- (axis cs:2, 89.95);
            \draw[gray, dashed, opacity=0.7] (axis cs:3, 86) -- (axis cs:3, 90.10);
            
        \end{axis}
    \end{tikzpicture}
    
    \caption{Impact of max iterations $T_{\max}$ (fix $m=3$) on MedQA-US (deepseek-v3.1).}
    \label{fig:ablation_t}
\end{figure}
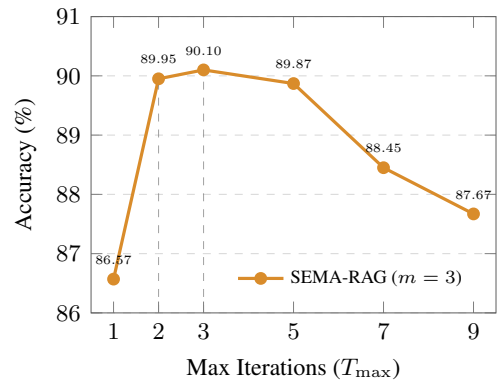

To configure this loop, we vary the maximum exploration depth $T_{\max}$ with $m=3$. 
Figure~\ref{fig:ablation_t} suggests that most of the benefit is captured within two rounds, with performance peaking around $T_{\max}\in\{2,3\}$; beyond that, deeper exploration saturates and can introduce noise. 
Notably, the two-round setting already outperforms our reproduced \textit{i}-MedRAG baseline (which uses three fixed retrieval rounds), suggesting that the gain comes from self-evolving, sufficiency-driven closed-loop exploration rather than simply increasing the number of iterations.

\subsection{Further Analysis}

\subsubsection{Synergy of the Multi-Agent Architecture}
Table~\ref{tab:ablation} shows that the full SEMA-RAG consistently performs best across both MedQA-US and PubMedQA*. 
Ablating any single agent leads to a clear drop, suggesting that the gains do not come from one isolated module but from role-specialized collaboration that mirrors the staged clinical workflow: I-Agent anchors clinically grounded interpretation, E-Agent drives sufficiency-driven evidence completion, and A-Agent consolidates and adjudicates evidence for option selection.

\subsubsection{Impact of Query Breadth}
Building on the depth analysis in Figure~\ref{fig:ablation_t}, we examine how the per-round query breadth $m$ affects E-Agent’s exploration. With $T_{\max}=2$ fixed, Table~\ref{tab:ablation_m} shows a clear monotonic trend: increasing $m$ improves accuracy, but the gains quickly taper. This suggests that expanding the query set helps cover complementary evidence gaps in early exploration, while additional branches beyond a moderate breadth tend to introduce overlapping or low-yield retrievals. We therefore use $T_{\max}=2$ and $m=3$ as the default setting in subsequent experiments, balancing coverage and efficiency.

\begin{table}[t]
    \centering
    \renewcommand{\arraystretch}{1.2}
    \small
    \setlength{\tabcolsep}{10pt}
    
    \begin{tabular}{lc}
        \toprule
        \textbf{Variant} & \textbf{MedQA-US (\%)} \\
        \midrule
        SEMA-m1 ($m=1$) & 86.72 \\
        SEMA-m2 ($m=2$) & 89.00 \\
        SEMA-m3 ($m=3$) & \textbf{89.95} \\
        \bottomrule
    \end{tabular}
    
    % 标题放在表格下方
    \caption{Effect of query breadth $m$ (fix $T_{\max}=2$) on MedQA-US (deepseek-v3.1).}
    \label{tab:ablation_m}
\end{table}

\subsection{Qualitative Case Study: A Head-to-Head Comparison}

Table~\ref{tab:case_study} shows a representative MedQA-US case comparing MedRAG and SEMA-RAG, where the decisive cue is the temporal constraint (hospital day 7), pointing to a hospital-acquired rather than community-acquired etiology.

\begin{table*}[t!]
\footnotesize
\setlength{\tabcolsep}{4pt} 
\renewcommand{\arraystretch}{1.1} 
\centering
\resizebox{\textwidth}{!}{%
\begin{tabularx}{\textwidth}{>{\raggedright\arraybackslash}X}
\toprule
\textbf{\makecell[c]{MedQA-US Question 0024}} \\
\midrule
\textbf{Context}: A 62-year-old patient has been hospitalized for a week due to a stroke. On hospital day 7, he develops a fever (38.4°C) and purulent cough. Vitals: HR 88, RR 20, BP 110/85. Physical exam: right basal crackles. Chest X-ray: new right-sided consolidation. \\
\textbf{Labs (Selected)}: WBC 8,900/mm³ (Neutrophils 72\%, Bands 4\%), Hb 16 g/dL, Platelets 280,000/mm³. \\
\textbf{Question}: What is the most likely causal microorganism? \\
\textbf{A.} \textit{Streptococcus pneumoniae} \quad \textbf{B.} \textit{Mycobacterium tuberculosis} \quad \textbf{C.} \textit{Haemophilus influenzae} \quad \textbf{D.} \textit{Staphylococcus aureus} \\
\midrule

% --- MedRAG Block (Light Gray) ---
\rowcolor[RGB]{242,242,242}
% 修改：使用 multicolumn 将这一行强制居中，同时保持 X 列宽度以填充背景色
\multicolumn{1}{>{\centering\arraybackslash}X}{\textbf{\textsc{Baseline: MedRAG}}} \\
\rowcolor[RGB]{250,250,250}
\textbf{Retrieved Evidence}: \\
"Community-acquired pneumonia (CAP) is most frequently caused by \textit{Streptococcus pneumoniae}..." ; "Stroke patients are at risk for aspiration pneumonia..." \\
\rowcolor[RGB]{250,250,250}
\textbf{Analysis \& Answer}: \\
The patient presents with typical signs of pneumonia. \sethlcolor{lightred}\hl{Given that \textit{S. pneumoniae} is the most common cause}, it is the likely pathogen... \\
\textbf{Prediction}: A \ \textcolor{red}{\xmark} \\
\midrule

% --- SEMA-RAG Block (Light Blue) ---
\rowcolor[RGB]{225,235,245}
% 修改：使用 multicolumn 将这一行强制居中
\multicolumn{1}{>{\centering\arraybackslash}X}{\textbf{\textsc{Ours: SEMA-RAG}}} \\

% Step 1: I-Agent
\rowcolor[RGB]{240,248,255}
\textbf{1. I-Agent (Interpretation)} \\
\rowcolor[RGB]{240,248,255}
\textit{Structured Schema}: \\
\textbf{Clinical Intent}: Infectious etiology \& pathogen identification \\
\textbf{Medical Entities}: stroke; HAP/aspiration pneumonia; right-basal crackles; new right consolidation; fever + purulent cough \\
\textbf{Clinical Constraints}: 62y; \sethlcolor{lightyellow}\hl{hospital day 7 }; post-stroke aspiration risk; neutrophil predominance \\
\textbf{Initial Query}: "hospital day 7 post-stroke pneumonia right consolidation most likely causative organism" \\

% Step 2: E-Agent
\rowcolor[RGB]{240,248,255}
\textbf{2. E-Agent (Exploration)} \\
\rowcolor[RGB]{240,248,255}
\textit{\textbf{Iteration 1}}: \\
\quad \textbf{Evidence}: "...Post-stroke patients have high risk of aspiration... pneumonia causes include anaerobes and streptococci..." \\
\quad \textbf{Gap}: \sethlcolor{lightyellow}\hl{Evidence does not distinguish pathogens based on hospitalization duration (day 7).} \\
\quad \textbf{Sufficiency}: $s_1=0$ (\textit{Insufficient}) $\rightarrow$ \textbf{Next Query}: "most likely pathogen hospital-acquired pneumonia vs community-acquired" \\
\rowcolor[RGB]{240,248,255}
\textit{\textbf{Iteration 2}}: \\
\quad \textbf{Evidence}: "...Hospital-acquired pneumonia (HAP) is defined as pneumonia $\ge$ 48h after admission..." \\
\quad \textbf{Key Find}: "For late-onset HAP ($\ge$ 5 days), \sethlcolor{lightyellow}\hl{common pathogens include \textit{Staphylococcus aureus} (MRSA) and \textit{Pseudomonas}..."} \\
\quad \textbf{Sufficiency}: $s_2=1$ (\textit{Sufficient}) \\

% Step 3: A-Agent
\rowcolor[RGB]{240,248,255}
\textbf{3. A-Agent (Adjudication)} \\
\rowcolor[RGB]{240,248,255}
\textbf{Report}: ... Hospital day 7 indicates HAP rather than CAP under standard definitions; combined with the candidate set, the evidence most consistently supports \textit{S. aureus}. 
\sethlcolor{lightyellow}\hl{Among the provided options, \textit{S. aureus} is the only matching HAP pathogen.} \\
\rowcolor[RGB]{240,248,255}
\textbf{Prediction}: D \ \greencheck \\
\bottomrule
\end{tabularx}
}
\caption{A case of how SEMA-RAG helps deepseek-v3.1 find the correct answer on MedQA-US (Question 0024) by making the \sethlcolor{lightyellow}\hl{key clinical constraint} explicit and retrieving \sethlcolor{lightyellow}\hl{decision-critical evidence}, while MedRAG’s single-round retrieval leads to a \sethlcolor{lightred}\hl{misleading rationale}.}
\label{tab:case_study}
\end{table*}

In the \textit{Baseline} block, MedRAG relies on a single round of static retrieval. Its evidence remains centered on generic pneumonia cues and aspiration risk, without explicitly anchoring retrieval to the hospital day 7 condition. This shifts the evidence toward typical community-acquired pathogens and leads to an incorrect selection of Streptococcus pneumoniae (Option A), which mismatches the inpatient, late-onset setting in the question.

In contrast, \textit{SEMA-RAG} makes the temporal constraint retrieval-actionable and carries it through to option selection. The \textbf{I-Agent} surfaces hospital day 7 as a key constraint, the \textbf{E-Agent} detects the missing distinction between community- and hospital-acquired spectra and performs targeted follow-up retrieval, and the \textbf{A-Agent} consolidates the resulting evidence and maps it to the candidate set, yielding the correct choice Staphylococcus aureus (Option D). Overall, the case shows how task decoupling with sufficiency-driven self-evolving retrieval helps form a more reliable evidence chain and prevents premature decisions under insufficient evidence.

\subsection{Cost and Efficiency Analysis}
Table~\ref{tab:efficiency_compact_deepseek} compares methods in terms of both accuracy and inference cost. Calls denotes the number of LLM invocations per question, while Retr. counts the number of vector retrieval operations; Time reports the average end-to-end latency per question; Tok./Q denotes the average total token consumption per question. Because SEMA-RAG employs sufficiency-driven early stopping, we report per-question averages over the MedQA-US set.

\begin{table}[H]
\centering
\small
\setlength{\tabcolsep}{5pt}
\begin{tabular}{lrrrrr}
\toprule
\textbf{Method} & \textbf{Calls} & \textbf{Retr.} & \textbf{Time} & \textbf{Acc.} & \textbf{Tok./Q} \\
& \textit{(\#)} & \textit{(\#)} & \textit{(s)} & \textit{(\%)} & \textit{(\#)} \\
\midrule
CoT & 1.0 & 0.0 & 2.5 & 77.53 & 713.7 \\
MedRAG & 1.0 & 1.0 & 3.2 & 77.14 & 2264.9 \\
\textit{i}-MedRAG & 3.0 & 9.0 & 8.8 & 74.78 & 21516.6 \\
\textbf{SEMA-RAG} & \textbf{4.8} & \textbf{3.4} & \textbf{9.5} & \textbf{89.95} & \textbf{19488.4} \\
\bottomrule
\end{tabular}
\caption{Efficiency comparison on MedQA-US (deepseek-v3.1).}
\label{tab:efficiency_compact_deepseek}
\end{table}

The results reveal three main patterns. First, SEMA-RAG consistently improves decision quality over single-round baselines. Second, these gains come with a moderate overhead relative to single-pass methods, as expected from multi-agent, multi-round inference. Third, compared with the iterative baseline \textit{i}-MedRAG, SEMA-RAG achieves a more favorable accuracy--efficiency trade-off, indicating that sufficiency-driven early stopping allocates additional computation more effectively than fixed-step iteration. Taken together, these results suggest that SEMA-RAG improves decision quality in a practically affordable regime, making the added overhead worthwhile for high-stakes medical QA.

\section{Related Work}

\subsection{Retrieval-Augmented Medical Reasoning}
Early medical LLMs systems largely depended on parametric knowledge or single-round retrieval \cite{luo_biogpt_2022,singhal_large_2023}. Methods such as MedRAG \cite{xiong-etal-2024-benchmarking} and MedCPT \cite{jin_medcpt_2023} improve domain retrieval with medical dual encoders and RAG pipelines, yet still follow a retrieve-once-then-answer pattern, which often falls short on questions requiring multi-hop evidence integration and clinically constrained reasoning \cite{jiang-etal-2023-active,trivedi-etal-2023-interleaving}.

Recent work has shifted from one-shot retrieval to iterative RAG \cite{gao_retrieval-augmented_2024,zhao_retrieval_2024}. In general domains, Self-RAG \cite{asai_self-rag_2023} and CRAG \cite{yan_corrective_2024} use self-reflection to trigger re-retrieval, while in healthcare \textit{i}-MedRAG \cite{xiong_improving_2025} iteratively refines queries via follow-up questions. However, without explicit clinical intent and constraints modeling, iterations may devolve into shallow rewrites, yielding inefficient retrieval, drift, and uncontrolled expansion \cite{zhao_retrieval_2024,zhu_large_2026}.

\subsection{Agentic Collaboration in Medicine}
Multi-agent systems enhance complex-task solving through role specialization and coordination \cite{10.24963/ijcai.2024/890,zong_triad_2024,tran_multi-agent_2025} (e.g., CAMEL \cite{li_camel_2023}, MetaGPT \cite{hong_metagpt_2024}). ReAct \cite{yao_react_2023} further couples reasoning with tool use, allowing agents to act and retrieve information during inference. In healthcare, MedAgents \cite{tang-etal-2024-medagents} and Agent-Hospital \cite{li_agent_2025} similarly show that multi-role clinical collaboration improves diagnosis and decision quality.

However, prior healthcare multi-agent work largely centers on deliberation under the assumption that key evidence is already in-context, leaving evidence acquisition unsystematic \cite{chen_evaluating_2025,gorenshtein_ai_2025,wang-etal-2025-survey}. In particular, gap identification, sufficiency-driven termination, and evidence adjudication/integration are often missing, weakening reliable external evidence grounding and closed evidence-chain formation in real clinical tasks \cite{li-2025-review,amugongo_retrieval_2025}.

\section{Conclusion}
We propose \textbf{SEMA-RAG}, a self-evolving multi-agent framework for medical question answering that restructures retrieval-augmented generation according to the staged process of clinical reasoning, with the \textbf{I-Agent} for clinical schema interpretation, the \textbf{E-Agent} for sufficiency-driven evidence exploration, and the \textbf{A-Agent} for evidence adjudication and final answer selection. Across five medical QA benchmarks and five LLM backbones, SEMA-RAG consistently outperforms strong baselines, improving the strongest baseline by an average of \textbf{+6.46} accuracy points, while ablations verify the necessity of the interpret--explore--adjudicate loop for reliable evidence-chain construction. Additional experiments further support its robustness across retrievers, smaller models, and more open-ended interactive settings. These findings suggest that medical RAG should move beyond static single-round retrieval toward more adaptive and reliable evidence construction.

\section*{Limitations}
Although we extend the evaluation beyond discrete-choice medical QA with additional open-ended and multi-turn benchmarks, the current study is still limited to benchmark-based settings rather than realistic clinical workflows such as longitudinal EHR reasoning or record-grounded decision support.

Our framework also depends on the quality and coverage of the retrieval corpus. If critical evidence is missing, outdated, or only partially retrieved, the self-evolving loop may still converge to incomplete grounding. In addition, the current sufficiency criterion is not explicitly designed for option-level separability or generative completeness.

Finally, SEMA-RAG introduces additional inference cost due to role specialization and multi-round exploration. Although this overhead is more efficient than fixed-step iterative baselines, it remains higher than single-round methods. The current design also lacks explicit relevance filtering during evidence accumulation, making performance sensitive to stopping criteria and exploration hyperparameters.

\bibliography{custom}

\appendix
\section*{Appendix}

\vspace{0.6em}

\section{Algorithm Pseudocode}
\label{app:algo}

Algorithm~\ref{alg:sema_rag} formally describes the inference flow of SEMA-RAG, highlighting the interaction between agents.

\begin{algorithm}[H]
\caption{Inference Process of SEMA-RAG}
\label{alg:sema_rag}
\begin{algorithmic}[1]
\REQUIRE Question $Q$, corpus $\mathcal{C}$, max iterations $T_{\max}$
\ENSURE Final answer $\tilde{y}$

\STATE \textbf{Stage 1: I-Agent}
\STATE $Q'=\langle o_{\mathrm{int}}, o_{\mathrm{ent}}, o_{\mathrm{cons}}, q_{\mathrm{init}}\rangle \leftarrow \mathrm{Agent}_{\mathrm{I}}(Pmt_{\mathrm{I}}, Q)$
\STATE $\hat{q}_{\mathrm{init}} \leftarrow \mathrm{Linearize}(Q')$
\STATE $C_0 \leftarrow \varnothing$
\STATE $\mathcal{Q}_1 \leftarrow \{\hat{q}_{\mathrm{init}}\}$

\STATE \textbf{Stage 2: E-Agent}
\FOR{$t = 1$ to $T_{\max}$}
    \STATE $\mathcal{D}_t \leftarrow \bigcup_{q \in \mathcal{Q}_t} \mathrm{TopK}(q)$
    \STATE $C_t \leftarrow \mathrm{Dedup}(C_{t-1} \cup \mathcal{D}_t)$
    \STATE $[s_t, g_t, \mathcal{Q}_{t+1}] \leftarrow \mathrm{Agent}_{\mathrm{E}}(Pmt_{\mathrm{E}}, [Q', \mathcal{Q}_t, C_t])$
    \IF{$s_t = 1$}
        \STATE \textbf{break}
    \ENDIF
    \IF{$\mathcal{Q}_{t+1}=\varnothing$}
        \STATE \textbf{break}
    \ENDIF
\ENDFOR
\STATE $T \leftarrow t$
\STATE $C^* \leftarrow C_t$

\STATE \textbf{Stage 3: A-Agent}
\STATE $\mathcal{S}^* \leftarrow \{(\mathrm{ID}(D_i), D_i)\mid D_i \in C^*\}$
\STATE $R \leftarrow \mathrm{Agent}_{\mathrm{A}}(Pmt_{\mathrm{adj}}, [Q, C^*, \mathcal{S}^*])$
\STATE $\tilde{y} \leftarrow \mathrm{Agent}_{\mathrm{A}}(Pmt_{\mathrm{ans}}, [Q, R])$
\RETURN $\tilde{y}$
\end{algorithmic}
\end{algorithm}

% ==========================================
% Section C: Dataset Details
% ==========================================
\section{Dataset Details}
\label{app:datasets}

We evaluate SEMA-RAG on five standard medical question-answering benchmarks from the MIRAGE suite: MMLU-Med, MedQA-US, MedMCQA, PubMedQA*, and BioASQ-Y/N. Table~\ref{tab:datasets} summarizes the dataset sizes and answer formats used in our experiments, and we briefly describe each benchmark below.

\subsection{MMLU-Med}
The MMLU-Med dataset is a subset of the Massive Multitask Language Understanding (MMLU) benchmark \cite{hendrycks_measuring_2021}. It covers six distinct subtasks: \textit{Clinical Knowledge}, \textit{Medical Genetics}, \textit{Anatomy}, \textit{Professional Medicine}, \textit{College Biology}, and \textit{College Medicine}. The questions are designed to measure knowledge acquired during preclinical and clinical medical training, formatted as 4-choice multiple-choice questions.

\subsection{MedQA-US}
MedQA-US \cite{jin_what_2020} is derived from the United States Medical Licensing Examination (USMLE). It represents highly complex clinical case studies that require multi-hop reasoning and domain-specific knowledge to solve. Following the standard setting in MIRAGE, we use the 4-option English version of the dataset. The questions typically present a patient vignette followed by a query about diagnosis, prognosis, or pharmacology.

\subsection{MedMCQA}
MedMCQA \cite{pal_medmcqa_2022} is a large-scale dataset collected from Indian medical entrance examinations (AIIMS and NEET-PG). It covers a wide range of 21 medical subjects, including surgery, pediatrics, and pharmacology. The questions vary significantly in difficulty and length, testing both memorization of medical facts and application of concepts in clinical scenarios.

\subsection{PubMedQA*}
PubMedQA \cite{jin_pubmedqa_2019} is a biomedical research question-answering dataset. The task requires answering "Yes", "No", or "Maybe" to a research question based on a provided abstract. \textbf{PubMedQA*} refers to the setting where the original context (abstract) is removed, forcing the model to retrieve external evidence to answer the question. This setting tests the system's ability to find relevant biomedical literature to support a scientific conclusion.

\subsection{BioASQ-Y/N}
BioASQ-Y/N is a subset of the BioASQ Task B benchmark \cite{tsatsaronis_overview_2015,krithara_bioasq-qa_2023}. It consists of biomedical questions that require a strict "Yes" or "No" answer. These questions are expert-constructed and reflect real-world information needs of biomedical researchers. The task is challenging because it often involves specific gene-disease associations or protein interactions that require precise fact-checking.

\begin{table}[t]
  \centering
  \begin{tabular}{lcl}
    \hline
    \textbf{Dataset} & \textbf{\#Samples} & \textbf{Task} \\
    \hline
    MMLU-Med   & 1089 & 4-choice MCQ       \\
    MedQA-US   & 1273 & 4-choice MCQ       \\
    MedMCQA    & 4183 & 4-choice MCQ       \\
    PubMedQA*  & 500  & 3-choice Y/N/M     \\
    BioASQ-Y/N & 618  & 2-choice Y/N       \\
    \hline
  \end{tabular}
  \caption{Statistics of the medical QA datasets from MIRAGE used in our experiments.}
  \label{tab:datasets}
\end{table}

% ==========================================
% Section D: Retrieval Corpus Details
% ==========================================
\section{Retrieval Corpus Details}
\label{app:corpus}

Following \textit{i}-MedRAG \cite{xiong_improving_2025} and MedRAG \cite{xiong-etal-2024-benchmarking}, we employ a hybrid retrieval corpus that combines medical textbooks with point-of-care clinical summaries, covering both foundational concepts and practical clinical knowledge.

\begin{description}
    \item[Textbooks] This component is the released medical textbook collection used in prior medical QA benchmarks \cite{app11146421}. It contains widely used reference textbooks spanning core biomedical sciences and clinical specialties, and is particularly helpful for queries requiring standard definitions, canonical mechanisms, and established medical principles.

    \item[StatPearls] StatPearls \cite{statpearls} is a point-of-care clinical review resource that provides high-yield summaries across diseases, diagnostics, and treatments. In our setup, we use the publicly available StatPearls articles (e.g., via the NCBI Bookshelf releases) as in prior work, which complements textbooks with concise, practice-oriented evidence for retrieval.
\end{description}

% ==========================================
% Section E: Error Analysis
% ==========================================
\section{Error Analysis}
\label{app:error}

To probe where SEMA-RAG can still fail, we analyze a representative MedQA-US error (Question 0060) in Table~\ref{tab:error_case_0060}. This case is informative because the system retrieves the correct biochemical cue at the \emph{class} level, yet still makes an incorrect discrete choice among the remaining candidates.

\begin{table}[t]
\centering
\small
\setlength{\tabcolsep}{5pt}
\renewcommand{\arraystretch}{1.15}
\begin{tabularx}{\linewidth}{>{\raggedright\arraybackslash}X}
\toprule
\textbf{\makecell[c]{Error Case (MedQA-US Q0060)}} \\
\midrule
\textbf{Question (brief)}: Septic shock with pelvic infectious focus; phenol/90$^\circ$C assay indicates a Lipid A--like motif $\Rightarrow$ Gram-negative signal. \\
\textbf{Options}: \\
\textbf{A.} Coagulase-positive, Gram-positive cocci \\
\textbf{B.} Encapsulated, Gram-negative coccobacilli \\
\textbf{C.} Spore-forming, Gram-positive bacilli \\
\textbf{D.} Lactose-fermenting, Gram-negative rods \\
\midrule

\rowcolor[RGB]{240,248,255}
\textbf{1) I-Agent}: Extracts anchors (pelvic source, shock/DIC-like labs) and treats the biochemical clue as the key discriminator. \\

\rowcolor[RGB]{240,248,255}
\textbf{2) E-Agent}: Retrieves evidence consistent with an LPS/Lipid A signature and correctly narrows the class to \textbf{Gram-negative}, ruling out \textbf{A/C} \ldots \\

\rowcolor[RGB]{255,245,245}
\textbf{3) A-Agent}: Commits among the remaining candidates without enforcing \emph{option-separating} evidence (rod vs.\ coccobacillus; lactose fermentation) \ldots \\
\textbf{Prediction}: B \ \textcolor{red}{\xmark} \qquad
\textbf{Ground Truth}: D \\
\bottomrule
\end{tabularx}
\caption{A representative failure where retrieval supports only \emph{class-level} elimination, while \emph{option-level} discrimination remains under-supported }
\label{tab:error_case_0060}
\end{table}

Here, \textbf{I-Agent} identifies the decision anchors from both presentation and assay. The clinical picture indicates severe sepsis with a pelvic infectious focus, accompanied by DIC-like abnormalities. The phenol-heating assay reveals a phosphorylated \textit{N}-acetylglucosamine dimer with multiple fatty acids, which strongly suggests a Lipid A–type structure. Guided by these anchors, \textbf{E-Agent} retrieves evidence linking Lipid A to LPS and therefore to Gram-negative organisms, which is sufficient to eliminate the Gram-positive distractors.

The failure arises when moving from class identification to option selection. The retrieved evidence supports ruling out the Gram-positive options, but it does not provide \emph{option-separating} signals within the remaining Gram-negative candidates. When the evidence report lacks an explicit bridge from the clinical setting to the discriminative phenotype expected in blood culture, \textbf{A-Agent} can be pulled toward salient surface descriptors and commit to an unsupported candidate.

This exposes a limitation of the current sufficiency and adjudication design. The loop may stop once it reaches a correct coarse conclusion, even though an additional round is still needed to uniquely determine the answer option. A practical implication is that sufficiency should be judged against \emph{option-level separability}, not only against class-level plausibility. A simple fix is to tighten E-Agent’s stopping rule by requiring evidence that supports one remaining option while directly excluding the other plausible candidates. If this condition is not met, the system should issue a final follow-up query targeting discriminative attributes and then re-adjudicate. We leave such option-aware sufficiency calibration to future work.

% ==========================================
% Section F: Implementation Details
% ==========================================
\section{Additional Implementation Details}
\label{app:implementation}

\paragraph{Computational Resources}
All experiments were executed in an API-based inference setting. The dense retrieval index was constructed and queried locally, while all LLM calls were served by the corresponding model providers. 

\paragraph{Retriever Settings}
We use the MedCPT Query Encoder and Article Encoder to embed queries and passages, respectively. We then perform FAISS-based dense retrieval over Textbooks and StatPearls and globally rank the retrieved candidates.

\section{Additional Robustness and Transfer Experiments}
\label{app:additional_exp}

\subsection{Retriever Robustness}
\label{app:retriever_robustness}

To examine whether SEMA-RAG depends strongly on a specific retriever, we further compare the domain-specific retriever MedCPT with the general-purpose retriever \textbf{qwen3-embedding-4b} \cite{zhang_qwen3_2025} on MedQA-US, using deepseek-v3.1 as the backbone. Table~\ref{tab:retriever_robustness} shows that SEMA-RAG consistently improves over the corresponding single-round retrieval baseline under both retrievers. Meanwhile, MedCPT remains the stronger default choice in this clinical setting, suggesting that domain-specific retrieval is still advantageous for medical QA.

\begin{table}[H]
\centering
\small
\setlength{\tabcolsep}{4pt}
\begin{tabular}{lcc}
\toprule
\textbf{Method} & \makecell[c]{\textbf{qwen3-}\\\textbf{embedding-4b}} & \textbf{MedCPT} \\
\midrule
CoT & 77.53 & 77.53 \\
MedRAG & 76.43 & 77.14 \\
SEMA-RAG & \textbf{87.43} & \textbf{89.95} \\
\bottomrule
\end{tabular}
\caption{Accuracy (\%) on MedQA-US with different retrievers (deepseek-v3.1).}
\label{tab:retriever_robustness}
\end{table}

\subsection{Smaller-Model Robustness}
\label{app:small_model_robustness}

To assess whether the gains of SEMA-RAG rely mainly on strong large-scale backbones, we further evaluate the framework on MedQA-US using a much smaller open-source model, \textbf{gemma3:4b} \cite{team_gemma_2025}. All other settings remain the same as in the main experiments. Table~\ref{tab:small_model_robustness} shows that although the absolute performance of all methods drops on the smaller model, SEMA-RAG still maintains a clear advantage over the strongest single-round baseline. This result suggests that task decoupling remains beneficial even when the base model has weaker instruction-following ability.

\begin{table}[H]
\centering
\small
\setlength{\tabcolsep}{8pt}
\begin{tabular}{l >{\centering\arraybackslash}p{1.8cm}}
\toprule
\textbf{Method} & \textbf{Acc. (\%)} \\
\midrule
CoT & 51.77 \\
MedRAG & 56.01 \\
SEMA-RAG & \textbf{60.41} \\
\bottomrule
\end{tabular}
\caption{Results on MedQA-US using gemma3:4b as the backbone.}
\label{tab:small_model_robustness}
\end{table}

\subsection{Beyond Discrete QA: Open-Ended and Interactive Settings}

To examine whether SEMA-RAG generalizes beyond discrete-choice medical QA, we further evaluate it on two benchmarks covering open-ended generation and multi-turn medical dialogue. In these experiments, we keep the I-Agent and E-Agent unchanged, and only adapt the final stage of the A-Agent to generate free-text outputs instead of selecting a discrete option. This setting isolates the contribution of the evidence loop under more open-ended output formats.

\paragraph{HealthBench.}
HealthBench is an open-ended health-domain benchmark designed to assess response quality under clinician-authored rubrics \cite{arora_healthbench_2025}. To test whether SEMA-RAG can transfer to grounded free-text generation with minimal modification, we evaluate deepseek-v3.1 on 500 randomly sampled English questions from HealthBench main using the official scoring script, and report the overall score together with three rubric dimensions: accuracy, completeness, and instruction following.

As shown in Table~\ref{tab:healthbench_transfer}, SEMA-RAG consistently outperforms MedRAG across all reported metrics. This suggests that the evidence completion loop remains effective when the target output shifts from option selection to grounded free-text generation, providing a stronger factual basis for open-ended responses.
\begin{table}[H]
\centering
\small
\setlength{\tabcolsep}{4pt}
\begin{tabular}{lcccc}
\toprule
\textbf{Method} & \textbf{Avg.} & \textbf{Acc.} & \textbf{Comp.} & \makecell[c]{\textbf{Instr.}\\\textbf{Follow.}} \\
\midrule
MedRAG   & 26.87 & 31.34 & 30.17 & 41.66 \\
SEMA-RAG & \textbf{33.64} & \textbf{37.22} & \textbf{38.30} & \textbf{47.58} \\
\bottomrule
\end{tabular}
\caption{Results on HealthBench for grounded open-ended response generation using deepseek-v3.1.}
\label{tab:healthbench_transfer}
\end{table} 

\paragraph{MAQuE.}
MAQuE is a multi-turn medical dialogue benchmark that evaluates response quality in simulated clinical communication settings \cite{gong_dialogue_2025}. Compared with single-turn QA, it places greater emphasis on maintaining robust, relevant, and contextually appropriate responses across iterative interactions.

To test whether SEMA-RAG remains effective in interactive settings, we evaluate deepseek-v3.1 on 200 randomly sampled MAQuE test cases using the official evaluation script, and report four communication-oriented metrics: Accuracy, Robustness, Relevance, and Empathy. Table~\ref{tab:maque_transfer} shows that SEMA-RAG maintains clear gains over MedRAG on all four metrics. This improvement suggests that sufficiency-driven exploration remains beneficial in interactive clinical scenarios, where the system must sustain grounded response generation over multiple turns.

\begin{table}[H]
\centering
\small
\setlength{\tabcolsep}{5pt}
\begin{tabular}{lcccc}
\toprule
\textbf{Method} & \textbf{Acc.} & \textbf{Rob.} & \textbf{Rel.} & \textbf{Emp.} \\
\midrule
MedRAG   & 52.50 & 64.86 & 74.00 & 66.40 \\
SEMA-RAG & \textbf{61.50} & \textbf{75.38} & \textbf{82.00} & \textbf{72.00} \\
\bottomrule
\end{tabular}
\caption{Results on MAQuE for multi-turn clinical response generation using deepseek-v3.1.}
\label{tab:maque_transfer}
\end{table}

Overall, these results suggest that the SEMA-RAG framework generalizes beyond discrete medical QA to more open-ended and interactive forms of clinical response generation.

\section{Prompt Templates}
\label{app:prompts}
To facilitate reproducibility, we provide the system instructions used for the three agents in SEMA-RAG. Note that the A-Agent uses two prompts for evidence adjudication and final answer selection, respectively.

\begin{figure*}[t]
\centering
\begin{tcolorbox}[width=0.96\textwidth]
{\scriptsize\ttfamily\linespread{1.02}\selectfont
I-Agent Prompt (Clinical Schema Interpreter).\\
\\
Role:\\
You are an expert clinician.\\
\\
Goal:\\
Given an unstructured medical question, extract an explicit Clinical Schema that makes the implied intent and constraints searchable. Focus on what must be retrieved and do not answer the question itself.\\
\\
Input:\\
Medical Question: \{research\_topic\}\\
\\
Task:\\
Identify:\\
1. clinical intent (task type),\\
2. core medical entities (salient concepts from the question),\\
3. key constraints (time course, demographics, setting, comorbidities, severity, contraindications, risk factors, anatomical or functional qualifiers),\\
4. a concise retrieval query aligned with the schema (q\_init).\\
\\
Key Instructions:\\
- Entities should be small, focused, and primarily grounded in the question itself.\\
- Include only the most retrieval-relevant concepts; avoid broad, redundant, or unnecessary enumeration.\\
- Merge obvious synonyms, near-duplicates, or simple morphological variants into one canonical medical expression when possible.\\
- Prefer the main clinical concept, condition, mechanism, finding, test, treatment, population, anatomical target, or other decision-critical concept that is necessary for retrieval.\\
- If the question provides candidate answers or options, do not mechanically include all of them as entities; include an option only if needed for retrieval or candidate discrimination.\\
- For multiple-choice, judgment, or open-ended questions, center the schema on the stem and its decision-critical medical concepts rather than listing answer choices.\\
- Constraints should capture only decision-relevant qualifiers explicitly stated or strongly implied by the question.\\
- Preserve key medical relations when they are essential for retrieval, such as derivation, origin, cause, association, indication, or contraindication.\\
- q\_init should retrieve the knowledge needed to answer the question, remain neutral, and avoid prematurely inferring a conclusion.\\
- q\_init should be short, medically precise, and should not simply concatenate all entities or options.\\
- Use precise medical terminology.\\
- Do not add explanations, rationale, or extra keys.\\
\\
Output JSON:\\
\{\\
\ \ "intent": "<short clinical task type>",\\
\ \ "entities": ["<entity1>", "<entity2>"],\\
\ \ "constraints": ["<constraint1>", "<constraint2>"],\\
\ \ "q\_init": "<one concise neutral search-style query>"\\
\}
}
\end{tcolorbox}
\caption{Prompt template for the I-Agent clinical schema interpreter.}
\label{fig:iagent_prompt}
\end{figure*}

\begin{figure*}[t]
\centering
\begin{tcolorbox}[width=0.96\textwidth]
{\scriptsize\ttfamily\linespread{1.02}\selectfont
E-Agent Prompt (Self-Evolving Explorer).\\
\\
Role:\\
You are an evidence sufficiency auditor and query refiner for medical question answering.\\
\\
Goal:\\
Determine whether the current retrieved evidence is sufficient to answer the medical question under the given Clinical Schema. Do not answer the question itself.\\
\\
Input:\\
Clinical Schema: \{clinical\_schema\}\\
Current Query Set: \{query\_list\}\\
Retrieved Evidence Summaries: \{summaries\}\\
\\
Key Instructions:\\
- Assess whether the current evidence sufficiently covers the key intent, entities, and constraints in the Clinical Schema.\\
- Judge sufficiency based on whether the evidence is enough to support final answer selection, or to distinguish among competing candidate answers when relevant.\\
- Evidence may be relevant yet still insufficient; do not mark sufficiency = 1 unless the evidence is adequate for confident answer selection.\\
- If the evidence is insufficient, identify the single most important missing fact, missing distinction, or unresolved clinical criterion.\\
- Generate 1 to 3 follow-up queries that directly target this gap.\\
- Follow-up queries must be specific, self-contained, non-redundant, and explicitly grounded in the Clinical Schema.\\
- For questions with candidate answers, prioritize queries that help distinguish among candidates rather than broad background expansion.\\
- Prefer targeted refinement over broad exploratory expansion.\\
- Do not repeat an existing query unless revision is necessary.\\
- If the current evidence is already sufficient, return no follow-up queries.\\
\\
Rules:\\
- If sufficiency = 1, set "gap" to "N/A" and "queries" to [].\\
- If sufficiency = 0, "gap" must be specific, concrete, and decision-relevant rather than generic.\\
- Queries should target missing clinical distinctions, time conditions, population constraints, contraindications, severity, mechanisms, diagnostic criteria, or option-level discrimination when relevant.\\
- Return JSON only.\\
\\
Output JSON:\\
\{\\
\ \ "sufficiency": 0 or 1,\\
\ \ "gap": "<short concrete description of the most important missing evidence>",\\
\ \ "queries": ["<query1>", "<query2>", "<query3>"]\\
\}
}
\end{tcolorbox}
\caption{Prompt template for the E-Agent self-evolving explorer.}
\label{fig:eagent_prompt}
\end{figure*}

\begin{figure*}[t]
\centering
\begin{tcolorbox}[width=0.96\textwidth]
{\scriptsize\ttfamily\linespread{1.02}\selectfont
A-Agent Prompt (Phase 1: Evidence Adjudicator).\\
\\
Role:\\
You are a medical evidence adjudicator.\\
\\
Goal:\\
Synthesize the final retrieved evidence into a concise, traceable report that can support final answer selection. Do not directly answer the question. Only organize, adjudicate, and summarize the evidence.\\
\\
Input:\\
Medical Question: \{research\_topic\}\\
Clinical Schema: \{clinical\_schema\}\\
Final Query Set: \{query\_list\}\\
Retrieved Evidence Summaries: \{summaries\}\\
\\
Key Instructions:\\
- Review the retrieved evidence in light of the medical question and Clinical Schema.\\
- Focus on the most decision-relevant evidence and remove redundancy.\\
- Identify which evidence directly supports a candidate conclusion, which evidence conflicts with it, and which evidence is only background, indirect, or weakly relevant.\\
- When multiple pieces of evidence overlap, merge them into one concise statement.\\
- When evidence is incomplete, uncertain, indirect, or conflicting, make that explicit rather than resolving it prematurely.\\
- Preserve traceability by attaching source identifiers or summary indices whenever available.\\
- Every claim in the report must be supported by the provided summaries; do not infer unsupported medical facts.\\
- Do not introduce external medical knowledge.\\
- Do not perform final answer selection.
}
\end{tcolorbox}
\caption{Prompt template for the A-Agent evidence adjudicator.}
\label{fig:aagent_phase1_prompt_a}
\end{figure*}

\begin{figure*}[t]
\centering
\begin{tcolorbox}[width=0.96\textwidth]
{\scriptsize\ttfamily\linespread{1.02}\selectfont
A-Agent Prompt (Phase 1: Evidence Adjudicator) (continued).\\
\\
Rules:\\
- Keep the report concise, traceable, and decision-oriented.\\
- Prefer evidence that is directly relevant to the question over general background knowledge.\\
- If there is no real conflicting evidence, return an empty list for "key\_conflicting\_or\_limiting\_evidence".\\
- If source identifiers are unavailable, use summary indices or short summary labels consistently.\\
- Do not repeat the same evidence across multiple fields unless necessary.\\
- Return JSON only.\\
\\
Output JSON:\\
\{\\
\ \ "question\_focus": "<one short sentence stating what must be decided>",\\
\ \ "key\_supporting\_evidence": [\\
\ \ \ \ \{\\
\ \ \ \ \ \ "claim": "<concise evidence-supported statement>",\\
\ \ \ \ \ \ "source\_ids": ["<source1>", "<source2>"]\\
\ \ \ \ \}\\
\ \ ],\\
\ \ "key\_conflicting\_or\_limiting\_evidence": [\\
\ \ \ \ \{\\
\ \ \ \ \ \ "claim": "<concise conflicting, uncertain, or limiting statement>",\\
\ \ \ \ \ \ "source\_ids": ["<source1>", "<source2>"]\\
\ \ \ \ \}\\
\ \ ],\\
\ \ "evidence\_synthesis": "<short integrated synthesis of what the evidence supports, what remains uncertain, and what distinction matters most for final answer selection>"\\
\}
}
\end{tcolorbox}
\caption{Prompt template for the A-Agent evidence adjudicator.}
\label{fig:aagent_phase1_prompt_b}
\end{figure*}

\begin{figure*}[t]
\centering
\begin{tcolorbox}[width=0.96\textwidth]
{\scriptsize\ttfamily\linespread{1.02}\selectfont
A-Agent Prompt (Phase 2: Evidence-Grounded Answering).\\
\\
Role:\\
You are a medical AI assistant.\\
\\
Goal:\\
Answer the multiple-choice medical question using the provided evidence adjudication report.\\
\\
Input:\\
Medical Question: \{research\_topic\}\\
Evidence Adjudication Report: \{adjudication\_report\}\\
\\
Key Instructions:\\
- Select exactly one final answer: A, B, C, or D.\\
- First rely on the evidence adjudication report.\\
- If the report contains relevant evidence, choose the option best supported by that evidence.\\
- If the report is incomplete, weak, or lacks directly relevant evidence, use medical knowledge to reason and choose the most appropriate answer.\\
- Do not output reasoning, JSON, code blocks, or any extra text.\\
\\
Output Format:\\
Final Answer: [A/B/C/D]
}
\end{tcolorbox}
\caption{Prompt template for the A-Agent evidence-grounded answerer.}
\label{fig:aagent_phase2_prompt}
\end{figure*}

\end{document}